\documentclass[lettersize,journal]{IEEEtran}
\usepackage{amsmath,amsfonts}
\usepackage{algorithmic}
\usepackage{algorithm}
\usepackage{array}
\usepackage[caption=false,font=normalsize,labelfont=sf,textfont=sf]{subfig}
\usepackage{textcomp}
\usepackage{stfloats}
\usepackage{url}
\usepackage{float}
\usepackage{verbatim}
\usepackage{graphicx}
\usepackage{cite}
\usepackage{multirow}
\usepackage{comment}

\begin{document}

\title{TreeGAN: Incorporating Class Hierarchy into Image Generation}

\author{Ruisi Zhang, Luntian Mou, Pengtao Xie
\thanks{R. Zhang, P. Xie are with the University of California, San Diego, CA, 92093, USA}
\thanks{L. Mou is with Beijing Key Laboratory of Traffic Engineering, Beijing University of Technology, China
}
\thanks{E-mail: ruz032@ucsd.edu, ltmou@pku.edu.cn, p1xie@ucsd.edu}
}

\markboth{Journal of \LaTeX\ Class Files,~Vol.~14, No.~8, August~2021}%
{Shell \MakeLowercase{\textit{et al.}}: A Sample Article Using IEEEtran.cls for IEEE Journals}


\maketitle

\begin{abstract}
Conditional image generation (CIG) is a widely studied problem in computer vision and machine learning. Given a class, CIG takes the name of this class as input and generates a set of images that belong to this class. In existing CIG works, for different classes, their corresponding images are generated independently, without considering the relationship among classes. In real-world applications,  classes are organized into a hierarchy and their hierarchical relationships are informative for generating high-fidelity images. In this paper, we aim to leverage class hierarchy for conditional image generation. We propose two ways of incorporating class hierarchy: prior control and post constraint. In prior control, we first encode the class hierarchy, then feed it as a prior into a conditional generator to generate images. We propose a novel hierarchy-aware embedding approach where similarities between embeddings are encouraged to be consistent with tree distances between classes, by minimizing log-determinant matrix divergence.
In post constraint, after  images are generated, we measure their consistency with the class hierarchy and use the consistency score to guide the training of the generator. To alleviate overfitting in training hierarchical classifiers, we propose a novel regularization approach -- self-supervised regularization, which uses a contrastive self-supervised learning (SSL) task to regularize the classification task, by jointly minimizing the SSL loss and classification loss.  
Experiments on various datasets demonstrate the effectiveness of our proposed methods.
\end{abstract}


\section{Introduction}
\IEEEPARstart{C}{onditional} image generation (CIG)~\cite{DBLP:journals/corr/MirzaO14} refers to the task of generating a set of images given a class label, where the generated images are aimed to be from this class. CIG has broad applications such as data augmentation~\cite{DBLP:conf/cvpr/BailoHS19,DBLP:conf/eccv/MilzRS18}, style transfer~\cite{DBLP:conf/accv/HobleyP18,DBLP:conf/aaai/WangTZSL0J20}, and image synthesis~\cite{DBLP:conf/cvpr/LiuLGWL19,DBLP:conf/aaai/Cao0WGS20,DBLP:conf/aaai/ChaGK19}, to name a few. In this work, we consider the problem of generating image sets given a collection of classes: for each class, generate a set of images belonging to this class. For example, given a set of animal classes including tiger, cat, snake, etc., we would like to generate a set of tiger images, a set of cat images, and so on. Existing CIG methods treat these classes as independent and generate a set of images for each class individually without considering their relationships.  
In real world applications, classes are typically organized into a hierarchy where children of a node represent the sub-classes of a class.  The hierarchical relationship between classes provides very valuable semantic clues for generating more realistic images. One clue could be: if we know class $A$ has smaller tree distance with $B$ than $C$, then the images of class $A$ should be more visually and semantically similar to $B$ than $C$. For example, in the oncology of animals, tiger is closer to cat than snake; therefore tiger images should be more similar to cat images than to snake images. 
Another clue could be: if class $A$ is a descendant of class $B$, then the images of class $A$ should inherit the visual and semantic attributes of  class $B$.  For instance, tiger and cat are both under the class of mammal; therefore the generated images for tiger and cat should share the attributes of mammal, such as having hair or fur. Such prior knowledge derived from class hierarchy should be incorporated into generative models to guide image generation. 

In this work, we aim to leverages class hierarchy for image generation and develop a generative model -- TreeGAN to achieve this goal. The input of TreeGAN is a hierarchy of classes and the output is a collection of image sets, one set of images for each class in the hierarchy. In TreeGAN, we propose two ways for incorporating class hierarchy: prior control and post constraint. These two ways can be used simultaneously. In prior control, the hierarchical relationship among classes is encoded and fed into a generative model to generate images. The generation process is guided by the encoded hierarchical relationship to produce images that are consistent with the class hierarchy. To learn hierarchy-aware class embeddings, we propose a novel embedding approach which encourages the similarities of class embeddings to be consistent with tree distances of classes, based on minimizing log-determinant matrix divergence~\cite{dhillon2008matrix}. In post constraint, after  images are generated, they are fed into a hierarchical classifier trained on real images in the class hierarchy to check whether the generated images are compatible with the class hierarchy. The compatibility score provides feedback to image generators to avoid generating hierarchy-incompatible images.  To alleviate overfitting in training hierarchical classifiers, we propose self-supervised regularization (SSR), a data-dependent regularization approach  based on self-supervised learning (SSL)~\cite{hadsell2006dimensionality,he2019moco,chen2020simple}. In SSR,    classification task and   SSL task are performed simultaneously. The SSL task is unsupervised, which is defined purely on input images without using any human-provided labels. Training a model using the SSL task can prevent the model from being overfitted to the limited number of class labels in the classification task.
Experiments on three datasets demonstrate the effectiveness of our methods.

\begin{figure}[ht!]
	\centering
	\includegraphics[width=\columnwidth]{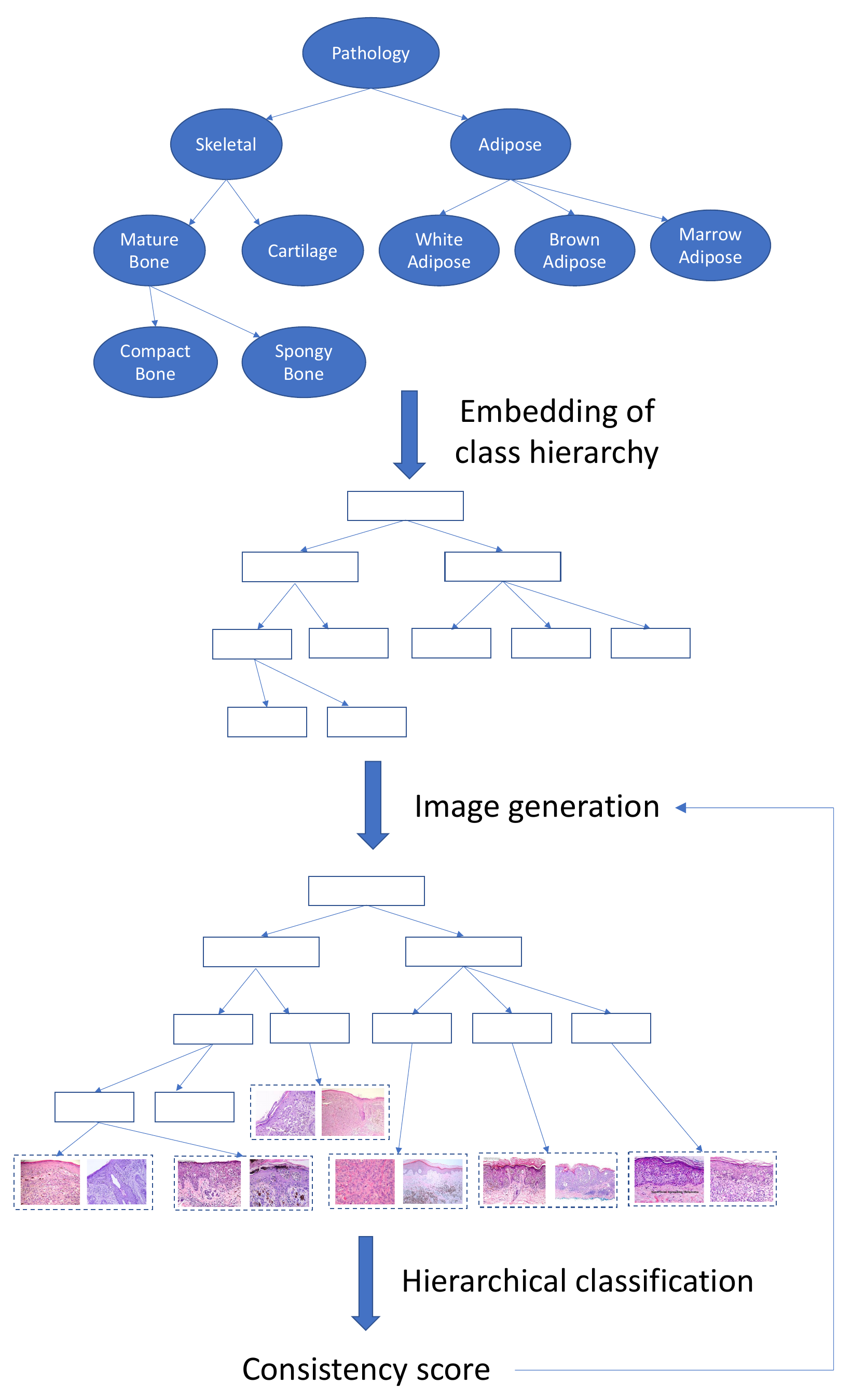}
	\caption{Illustration of TreeGAN. Given a class hierarchy, we first leverage this hierarchy to learn embeddings for each class in the hierarchy. These embeddings incorporate the hierarchical relationship among classes. For each class on the leaf node in the hierarchy, we feed its embedding into a conditional GAN to generate images belonging to this class. Given the generated images, we feed them into an offline-trained hierarchical classifier to check whether the generated images are compatible with the class hierarchy. The consistency score is used to guide the image generator to generate hierarchy-compatible images.}  
	\label{fig1}
\end{figure}

The major contributions of this paper are as follows.
\begin{itemize}
    \item We propose TreeGAN, a deep generative model which incorporates class hierarchy to generate images.
    \item We propose two ways of incorporating class hierarchy: prior control and post constraint.
    \item We propose a novel hierarchy-aware class embedding approach which achieves consistency between embedding similarities and tree distances by minimizing log-determinant matrix divergence.
    \item We propose self-supervised regularization, which is a data-dependent regularizer based on SSL, to reduce the risk that the image encoder is biased to the data-deficient classification task on  small-sized training data.
    \item We demonstrate the effectiveness of TreeGAN on various datasets. 
\end{itemize}
The rest of the paper is organized as follows. Section 2 reviews related works. Section 3 and 4 present the method and experimental results.  Section 5 concludes the paper and discusses future works. 

\section{Related Works}
\subsection{Conditional Image Generation}
Generating images conditioned on class names or other types of texts have been widely studied. Mansimov \textit{et al.} (2016) proposed an encoder-decoder architecture for text-to-image generation. The encoder of text and the decoder of image are both based on recurrent networks. Attention is used between image patches and words. StackGAN~\cite{zhang2017stackgan} first uses a GAN to generate low-resolution images, which are then fed into another GAN to generate high-resolution images. AttnGAN~\cite{xu2018attngan} synthesizes fine-grained details at different subregions of  images by paying attention to  relevant words in  natural language description. DM-GAN~\cite{zhu2019dm} uses a dynamic memory module to refine fuzzy image contents when the initial images are not well generated and designs a memory writing gate to select  important text information. Obj-GAN~\cite{Li_2019_CVPR} proposes an object-driven attentive image generator to synthesize salient objects by paying attention to the most relevant words in  text description and  pre-generated semantic layout. MirrorGAN~\cite{qiao2019mirrorgan} uses an autoencoder architecture, which generates an image from a text, then reconstructs the text from the image. We summarized the details of different conditional image generation methods in Table~\ref{tab:congen}.

\begin{table*}[htbp]
	\centering
	\renewcommand{\arraystretch}{1.2}
	\caption{Comparison of Conditional Image Generation Methods}
	\label{tab:congen}
	\begin{tabular}{p{2.5cm} | p{3.5cm} | p{3.4cm} | c | p{4.5cm} }
		\hline
		Methods & Text Encoder & Image Encoder & Image Size & Image Generation Approach  \\
		\hline
		GAN-INT-CLS\cite{reed2016generative} &  character level CNN or LSTM & deep CNN & 64 $\times$ 64 & train a deep convolutional GAN conditioned on text features\\\hline
		StackGAN~\cite{zhang2017stackgan} &character level CNN or LSTM & deep CNN&  256 $\times$ 256 & generating high-resolution photo-realistic image with multi generators and multi discriminators organized in a tree-like structure\\\hline
		AttnGAN~\cite{xu2018attngan}& bi-directional LSTM & Inception-v3 pretrained on ImageNet&  256 $\times$ 256 & use attention-driven, multi-stage refinement for fine-grained  generation\\\hline
		DM-GAN~\cite{zhu2019dm}& transform text description to a sentence feature and several word features  & Inception-v3 pretrained on ImageNet &  256 $\times$ 256 &  introduce a dynamic memory module to refine fuzzy image contents, when the initial images are not well generated\\\hline
		Obj-GAN~\cite{Li_2019_CVPR}& deep attention multi-modal similarity model& CNN mapping images to semantic vectors&  256 $\times$ 256 &  synthesize  complex scenes by paying attention to the most relevant words in the text description and the pre-generated semantic layout\\\hline
		MirrorGAN~\cite{qiao2019mirrorgan}& semantic text embedding to transform text into local word-level features and global sentence-level features&Inception-v3 pretrained on ImageNet &  256 $\times$ 256 &  learning image generation by redescription \\
		\hline
	\end{tabular}
\end{table*}

\subsection{Self-supervised Learning}

Self-supervised learning has been widely applied to other application domains, such as computer vision, where  various strategies have been proposed to construct auxiliary tasks, based on temporal correspondence~\cite{li2019joint, wang2019learning}, cross-modal consistency~\cite{wang2019reinforced},  rotation prediction ~\cite{gidaris2018unsupervised, sun19ttt}, image inpainting~\cite{pathak2016context}, automatic colorization~\cite{zhang2016colorful}, context prediction~\cite{nathan2018improvements}, etc.
Some recent works studied self-supervised representation learning based on instance discrimination~\cite{wu2018unsupervised} with contrastive learning. \cite{oord2018representation}  proposed contrastive predictive coding (CPC), which  predicts the future in latent space by using powerful autoregressive models, to extract useful representations from high-dimensional data.  
\cite{bachman2019learning}  proposed a self-supervised representation learning approach based on maximizing mutual information between features extracted from multiple views of a shared context. 

MoCo~\cite{he2019moco}  and SimClr~\cite{chen2020simple} learn image encoders by predicting whether two augmented images are created from the same original image. 
\cite{srinivas2020curl} proposed to learn  contrastive unsupervised representations for reinforcement learning. \cite{khosla2020supervised} investigated supervised contrastive learning, where clusters of points belonging to the same class were pulled together in the embedding space, while clusters of samples from different classes were pushed apart. \cite{klein2020contrastive} proposed a contrastive self-supervised learning approach for commonsense reasoning.

\section{Methods}

\begin{figure}[t]
	\centering
	\includegraphics[width=0.9\columnwidth]{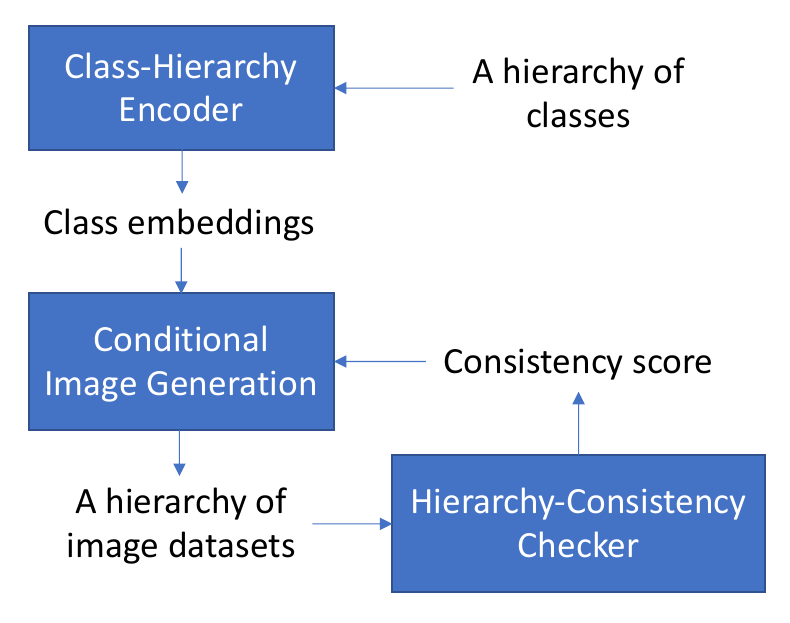}
	\caption{Illustration of TreeGAN. 
	}  
	\label{fig2}
\end{figure}

In this section, we introduce the proposed TreeGAN which takes a hierarchy of classes as input and generates an image set for each class. The detailed illustration is shown in Figure~\ref{fig1}. During the generation process, the hierarchical relationship among classes is leveraged. TreeGAN is composed of three modules: a class-hierarchy encoder (CHE), a conditional image generator (CIG), and a hierarchy-consistency checker (HCC). The three modules are trained jointly end-to-end.   We propose two ways to leverage  class hierarchy for image generation: prior control and post constraint. Prior control is imposed before  images are generated. Post constraint is used after  images are generated. In prior control, we use the class-hierarchy encoder to learn embeddings for the hierarchy of classes where the embedding of each class captures the hierarchical relationship between this class and other classes. Then the hierarchy-aware class embeddings are fed into the conditional image generator to generate a set of images for each class. In post constraint, we first train a hierarchical classifier offline, which is used as the hierarchy-consistency checker (HCC). After the hierarchy of image sets are generated, we feed them into the HCC to check whether the generated images are compatible with the class hierarchy. If these images can be classified correctly by the classifier, then they are highly compatible with the hierarchy. The consistency score is used to guide the generator to generate hierarchy-compatible images. Detailed illustration is shown in Figure~\ref{fig2}. In the sequel, we introduce each of these modules.



\subsection{Class-Hierarchy Encoder}

Given a hierarchy of classes where each class has a textual name, we use the class-hierarchy encoder (CHE) to learn embeddings of these classes where the embeddings simultaneously capture the hierarchical relationship among classes and the semantics of each class. We learn the embeddings based on the following principle: the similarities of class embeddings should be consistent with the tree distances of classes, i.e., if two classes have small tree distance, their embeddings should be similar. The tree distance of two classes is the length of the path connecting these two classes in the tree. A smaller tree distance indicates that the two classes are close to each other in the hierarchy. 
Let $S\in\mathbb{R}^{K\times K}$ be a matrix where $K$ is the number of classes in the hierarchy. $S_{ij}=\textrm{sim}(\theta_i, \theta_j)$, where $\theta_i$ and $\theta_j$  are the embeddings of class $i$ and $j$ respectively. $\textrm{sim}(\cdot,\cdot)$ denotes the cosine similarity between two embedding vectors. Let $T\in\mathbb{R}^{K\times K}$ be another matrix. $T_{ij}=1/(d+1)$ where $d$ denotes tree distance between class $i$ and $j$. $S$ and $T$ are normalized using min-max normalization. Both $S$ and $T$ measure the similarity between classes, but from different perspectives: $S$ measures class similarity based on how class embeddings are similar to each other; $T$ measures class similarity based on the length of paths connecting classes in the tree. We would like $S$ and $T$ to be consistent, i.e., if two classes have large similarity in $T$, they should have large similarity in $S$ as well. To achieve this goal, we encourage the log-determinant divergence (LDD)~\cite{dhillon2008matrix} between $S$ and $T$ to be small. LDD is a type of Bregman matrix divergence~\cite{dhillon2008matrix} which measures the ``nearness" between matrices. A smaller LDD indicates that two matrices are more close to each other. The LDD between $S$ and $T$ is defined as follows:
\begin{equation}
\begin{aligned}
    \textrm{tr}(ST^-)-\textrm{logdet}(ST^-)
\end{aligned}
\end{equation}
where $\textrm{tr}(\cdot)$ denotes matrix trace and $\textrm{logdet}$ denotes the log of matrix determinant. Previous embedding methods~\cite{pbg,glove} only consider local and low-order hierarchical relationships between child-parent nodes while our method is able to incorporate long-range hierarchical relationships between any pair of nodes on the tree. 

\subsection{Conditional Image Generator}
Given class embeddings which capture the hierarchical relationship among classes, we feed the embedding of each class into a conditional image generator (CIG) to generate images belonging to this class. Any CIG, such as those proposed in \cite{reed2016generative,Han17stackgan2,LeicaGAN,tao2020df}, is applicable in our framework. 
The focus of this paper is to incorporate class hierarchy into generative models instead of improving generative models themselves. We do not attempt to make novel contributions in this module. 


\subsection{Hierarchy-Consistency Checker}
\label{sec:hcc}
Given the images generated by CIG, the hierarchy-consistency checker checks whether the generated images are consistent with the class hierarchy. Given the training data which contains a hierarchy of classes and each class has a set of real images, we train a hierarchical classification model on these images offline. After training, we use the hierarchical classifier to measure the consistency between generated images and class hierarchy. Given the hierarchy of images generated by the conditional image generator, we feed them into the  hierarchical classifier to measure  classification errors. Small classification errors indicate that the generated images are more compatible with the class hierarchy. The hierarchical classifier has multiple levels of classifiers. At each level in the hierarchy, there is a multi-class classifier distinguishing  classes at this level. Given a generated image $X_y$ where $y$ denotes the class from which this image is generated, let $f_i^{(k)}(X_y)$ denote the prediction probability that the input image belongs to the $i$-th class at the $k$-th level. The classification loss is defined as
\begin{equation}\label{bwithg-loss}
\begin{aligned} 
b(X_y)=\sum_{k=1}^{K}- \log \left(\frac{e^{f^{(k)}_{a_{k}(y)}(X_y)}}{\sum_{j=1}^{M_k} e^{f_{j}^{(k)}(X_y)}}\right)
\end{aligned}
\end{equation}
where $K$ is the number of levels in the class hierarchy, $a_k(y)$ is the ancestor of class $y$ at the $k$-th level, and $M_k$ is the number of classes at the $k$-th level. A smaller loss indicates that $X_y$ is more compatible with the class hierarchy. Note that $X_y$ is a function of the weight parameters of the generator. To encourage the generator to generate hierarchy-compatible images, we train its weight parameters by minimizing the classification loss on each generated image. 

When the amount of training data for the hierarchical classifier is limited, there is a significant risk that the classifier is overfitted to the training data and generalizes less well on ``test" data which are unseen images generated by the CIG. To alleviate overfitting, we propose  self-supervised regularization (SSR), a data-dependent regularization approach which uses a self-supervised learning task to regularize the hierarchical classification task. Self-supervised learning (SSL)~\cite{hadsell2006dimensionality,he2019moco,chen2020simple} is an unsupervised learning approach which defines auxiliary tasks on input data without using any human-provided labels and learns data representations by solving these auxiliary tasks. In existing SSL approaches, an SSL task and a target task are performed sequentially. An image encoder is first trained by solving the SSL task defined on a large collection of unlabeled images. Then this encoder is  finetuned by solving the target task. A potential drawback of performing the SSL task and target task sequentially is that the image encoder learned in the SSL task may be overridden after being finetuned in the target task. If training data in the target task is small, the finetuned encoder has a high risk of being overfitted to the training data.  To address this problem,  we propose to perform the SSL task and the target task (which is hierarchical classification) simultaneously by jointly optimizing the SSL loss and classification loss where the SSL loss serves as a regularization term.   SSR  enforces the image encoder to jointly solve two tasks: an unsupervised SSL task and a supervised hierarchical classification task. Due to the presence of the SSL task, the model is less likely to be biased to the classification task defined on the small-sized training data. We use MoCo~\cite{he2019moco} as the SSL task. It creates augmentations of original data examples and defines an auxiliary task which judges whether two augmented data examples originate from the same original data example.  

\subsection{Loss Function}

Putting these pieces together, we are ready to define the overall loss function. Let $\mathcal{H}$ denote the class hierarchy, $\mathcal{C}$ denote all the classes (including those on the internal nodes and leaf nodes) in the hierarchy, and $\mathcal{L}$ denote the set of classes on the leaf nodes of $\mathcal{H}$.  Let $\boldmath{e}_c$ denote the embedding of class $c$, $\mathcal{E}=\{\boldmath{e}_c|c\in \mathcal{C}\}$, and $f(\mathcal{E}, \mathcal{H})$ be the loss of learning hierarchical embeddings. Let $g(G,D,\boldmath{e}_c)$ denote the GAN loss of generating images for class $c$, where $G$ is the generator and $D$ is the discriminator. All the classes share the same generator and discriminator. We define the hierarchical classification loss on generated images in class $c$ as $h(G,c)=\sum_{X_c\in \mathcal{G}(G,c)} b(X_c)$ where $\mathcal{G}(G,c)$ are all generated images in class $c$ and $b(X_c)$ is given in Eq.(\ref{bwithg-loss}). The overall loss is defined as follows:

\begin{equation}
\label{eq:overall}
    \textrm{min}_{D,\mathcal{E}} \; \textrm{max}_{G} \;  \sum_{c\in \mathcal{L}}  (g(G,D,e_c) - \lambda_1 h(G, c)) + \lambda_2 f(\mathcal{E}, \mathcal{H})
\end{equation}
where $\lambda_1$ and $\lambda_2$ are regularization parameters. The generator aims to generate images that are indistinguishable from the real images and consistent with the class hierarchy.

\section{Experiments}
In this section, we present experimental results on three datasets. Each dataset has a hierarchy of classes and each class has a set of real images. 

\subsection{Dataset}

Three datasets are used in the experiments: Animal, Vegetable, and Pathology. The Animal dataset has 15 classes, which are organized into a hierarchy of three levels. There are 10 classes on the leaf nodes, which are otter, bear, fox, wolf, dog, cat, lion, tiger, raccoon and panda. Each leaf class has 500 images collected from ImageNet~\cite{DBLP:conf/cvpr/DengDSLL009} and Google Open Image Dataset\footnote{\url{https://storage.googleapis.com/openimages/web/index.html}}. The Vegetable dataset has 7 classes, which are organized into a hierarchy of three levels. There are 5 classes on the leaf nodes, which are broccoli, bok choy, mushroom, pumpkin, and cucumber. Each leaf class has 500 images collected from ImageNet. The Pathology dataset has 8 classes, which are organized into a hierarchy of three levels. There are 6 classes on the leaf nodes, which are  eosinophil,  lung,  lymphocyte, monocyte, neutrophil, and retina. Each leaf class has 500 images collected from cell dataset~\cite{cell_dataset}, lung dataset~\cite{lung_dataset} and retina dataset~\cite{retina_dataset}.

\begin{table*}[t]
    \caption{ IS (larger is better) and FID (smaller is better) scores achieved by different methods on three datasets.}
	\centering
\begin{tabular}{l|c|c|c|c|c|c}
	\hline
	& \multicolumn{2}{c|} {Animal} 	& \multicolumn{2}{c|} {Vegetable}
	& \multicolumn{2}{c} {Pathology}\\
	\hline
 & IS$\uparrow$ & FID$\downarrow$& IS$\uparrow$ & FID$\downarrow$& IS$\uparrow$ & FID$\downarrow$\\
 \hline
 DF-GAN & 3.30& 72.49& 3.97 & 131.13& 2.41& 127.66\\
  DF-GAN + Ours & \textbf{3.75}& \textbf{61.53}&\textbf{4.30} & \textbf{109.38}& \textbf{2.49}& \textbf{124.05}\\
  \hline
   LeicaGAN & 3.63&92.60 & 4.01& \textbf{128.47}& 2.12& 186.84\\
  LeicaGAN + Ours & \textbf{4.92}&\textbf{89.23} &\textbf{5.54} &136.25 &\textbf{3.10} &\textbf{179.67} \\
 \hline
    StackGAN++  &3.70 &106.52 & 4.11&154.68 & 2.48& 195.88\\
  StackGAN++ + Ours &\textbf{3.90} & \textbf{101.59}& \textbf{4.33}& \textbf{145.44}& \textbf{3.10}& \textbf{186.45}\\
  \hline
      GAN-INT-CLS  & 3.66&87.49 &\textbf{4.04} &129.54 &2.15 &88.01 \\
  GAN-INT-CLS + Ours &\textbf{3.87} &\textbf{85.82} & 4.03&\textbf{112.76} &\textbf{2.23} & \textbf{85.86}\\
 \hline
\end{tabular}
 \label{tb:all}
\end{table*}

\subsection{Experimental Settings}
In Eq.(\ref{eq:overall}), $\lambda_1$ was set to 0.1 and $\lambda_2$ was set to 0.01. When training our proposed hierarchy-aware class-embeddings, an SGD optimizer was used with a learning rate of 0.1. The embedding size was set to 256 when our methods were applied to  DF-GAN and  LeicaGAN, and set to 1024 when applied to StackGAN++ and GAN-INT-CLS. 
When training hierarchical classifiers with self-supervised regularization, an SGD optimizer was used with a learning rate of 0.03. When training image generators, 
the mini-batch size was set to 20 when our methods were applied to  DF-GAN, LeicaGAN and StackGAN++ and set to 64 when applied to  GAN-INT-CLS.

Most hyperparameter settings follow those in  \cite{reed2016generative}, \cite{Han17stackgan2}, \cite{LeicaGAN} and \cite{tao2020df}. We tuned the epoch and batch size during the training process to get better image generation results. In experiments based on Df-GAN, LeicaGAN, and StackGAN++, we initially set the training epochs to 5000 and evaluated the generated images at every 100 epochs after 1000.  We found that for most of the datasets, the IS/FID scores converge at about 2000 epochs in experiments based on DF-GAN and LeicaGAN. Therefore, we set the epochs in these experiments to 2000.  For most of the datasets, the IS/FID scores converge at about 3000 epochs in experiments based on StackGAN++. Therefore, we set the epochs in these experiments to 3000. 
In experiments based on GAN-INT-CLS, we initially set the training epochs to 10000 and evaluated the generated images at every 1000 epochs. For most of the datasets, the IS/FID scores converge at about 6000-7000 epochs. Therefore, we set the epochs these experiments to 7000.

When generating images, we followed the setting in \cite{reed2016generative} and set the batch size to 64. We also tuned the batch size in 32 and 128, but the performance did not improve greatly. When generating 256x256 images, we used a smaller batch size to ensure the model can fit into the GPU memory. We tried a batch-size of 20 and 10. The performance is  similar and the training time was reduced when the batch size was set to 20.

When training  hierarchical classifiers on $256 \times 256$ images, we used 80\% of the dataset for training, 10\% for validation, and 10\% for testing. 
During training, the loss started to converge at about 70-100 epoch for different datasets. 
When training hierarchical classifier on $64 \times 64$ images, we used 80\% of the dataset for training, 10\% for validation, and 10\% for testing. 
During training, the loss started to converge at about 110-160 epochs for different datasets. 

\begin{table*}[ht]
\centering
\begin{center}
\caption{\label{tab:hp:hcc-acc2}
 Hierarchical classification accuracy (Acc) on training set, validation (Val) set and test set. The last column shows the difference between training accuracy and test accuracy. A smaller difference indicates less overfitting. 
}
\begin{tabular}{c|c|c|c|c|c}
	\hline
    Dataset(Image Size) &  Methods & Train Acc& Val Acc & Test Acc & Train\_{Acc} - Test\_{Acc}\\
	\hline
	Animal (64 $\times$ 64)& SSR  &  76.90\% & 78.00\% & 73.40\% & \textbf{3.50}\%\\
	Animal (64 $\times$ 64)& No SSR  & 83.88\%  & 82.40\% & 71.60\% & 12.28\%\\
	\hline
	Animal (256 $\times$ 256)& SSR  & 73.88\% & 73.20\% & 65.80\% & 8.08\%\\
	Animal (256 $\times$ 256)& No SSR  & 79.10\% & 77.80\% & 74.40\% & \textbf{4.70}\% \\
		\hline
	Vegetable (64 $\times$ 64)& SSR & 84.75\% & 83.60\%  &79.60\% & \textbf{5.15}\%\\
	Vegetable (64 $\times$ 64)& No SSR & 91.25\% & 87.40\% & 83.20\% & 8.05\%\\
		\hline
	Vegetable (256 $\times$ 256)& SSR & 87.80\% & 85.20\% & 79.20\% & 8.60\%\\
	Vegetable (256 $\times$ 256)& No SSR & 85.70\% & 86.20\% & 81.20\% & \textbf{4.50}\%\\
		\hline
	Pathology (64 $\times$ 64)& SSR & 84.08\%  & 81.33\%& 77.67\% & \textbf{6.41}\%\\
	Pathology (64 $\times$ 64)& No SSR & 88.58\%  & 89.33\% & 80.67\% & 7.91\%\\
		\hline
	Pathology (256 $\times$ 256)& SSR & 89.96\% & 86.67\% & 81.67\% & \textbf{8.29}\%\\
	Pathology (256 $\times$ 256)& No SSR & 90.67\% & 88.67\% & 80.33\% & 10.34\%\\
	\hline
\end{tabular}
\end{center}
\end{table*}

\subsection{Hardware and software}
Our experiments ran on NVIDIA tesla V100 GPUs, and the networks were trained using PyTorch. On average, it takes about 8 hours to train for one class in experiments based on Df-GAN, 6 hours to train for one class in experiments based on LeicaGAN, 5 hours to train for one class in experiments based on StackGAN++, and 4 hours to train for one class in experiments based on GAN-INT-CLS. The training of hierarchical classifiers is within two hours.

\subsection{Evaluation metrics}
We use two metrics to evaluate the quality of generated images: Fréchet Inception Distance and Inception Score.

\subsubsection{Fréchet Inception Distance} 

The Fréchet Inception Distance (FID)~\cite{heusel2017gans} is a measure of similarity between two datasets of images. It was shown to correlate well with human judgement of visual quality and is widely used to evaluate the quality of generated images by Generative Adversarial Networks. FID is calculated by computing the Fréchet distance between two Gaussians fitted to feature representations of the Inception network. Given two Gaussians with mean vector $m$ and variance matrix $C$: $(m_1, C_1)$ and $(m_2, C_2)$, the FID is given by

\begin{equation}
\begin{array}{l}
    d((m_1,C_1),(m_2,C_2))=\|m_1-m_2\|_2^2+\text{Tr}(C_1+C_2-2(C_1C_2)^{1/2})
\end{array}
\end{equation}

\subsubsection{Inception Score.} 

The Inception Score~(IS)~\cite{salimans2016improved} is a metric for automatically evaluating the quality of image generation models. The IS uses an Inception-v3 network pre-trained on ImageNet to calculate a statistic of the network’s outputs when applied to generated images
\begin{equation}
    IS(G) = \exp(\mathbb{E}_{x\sim p_g}D_{KL}(p(y|x)||p(y)))
\end{equation}
where $x \sim p_g$ indicates that $x$ is an image sampled from $p_g$,
$D_{KL}(p||q))$ is the KL-divergence between the distributions
$p$ and $q$, $p(y|x)$ is the conditional class distribution, and  $p(y)= \int_xp(y|x)p_g(x)$
is the marginal class distribution.

\begin{table*}[t]
    \caption{ IS (larger is better) and FID (smaller is better) scores achieved by different ablation settings on three datasets.}
	\centering
\begin{tabular}{l|l|c|c|c|c|c|c}
	\hline
\multirow{2}{*}{Base Model} &\multirow{2}{*}{Ablation Settings}	& \multicolumn{2}{|c|} {Animal} 	& \multicolumn{2}{|c|} {Vegetable}
	& \multicolumn{2}{|c} {Pathology}\\
	\cline{3-8}
& & IS$\uparrow$ & FID$\downarrow$& IS$\uparrow$ & FID$\downarrow$& IS$\uparrow$ & FID$\downarrow$\\
\hline
    \multirow{4}{*}{DF-GAN} & TreeGAN & \textbf{3.75} & \textbf{61.53} & \textbf{4.30} & \textbf{109.38} & \textbf{2.49} & \textbf{124.05}\\
                            & No-PC   & 3.55 & 65.94 & 4.09 & 123.64& 2.42 & 126.60\\
                            & No-SSR  & 3.48 & 82.76 & 3.42 & 180.44& 2.38 & 131.17\\ 
                            & PBG     & 3.63 & 85.02 & 3.68 & 138.43& 2.34 & 125.63\\
                            & GloVe   & 3.58 & 92.18 & 3.79 & 163.48& 2.41 & 127.58\\
                            & SEG     & 3.53 & 69.57 & 3.92 & 116.41 & 2.37 & 128.11\\
\hline
 \multirow{4}{*}{LeicaGAN} & TreeGAN& \textbf{4.92} & \textbf{89.23} & \textbf{5.54} & \textbf{136.25} & \textbf{3.10} & \textbf{179.67}\\
                           & No-PC  & 4.91 & 94.57  & 5.30 & 143.53 & 2.96 & 192.86\\
                           & No-SSR & 4.69 & 99.26  & 5.23 & 169.49 & 2.98 & 199.48\\ 
                           & PBG    & 4.65 & 101.90 & 5.53 & 155.39 & 2.98 & 188.27\\
                           & GloVe  & 4.71 & 102.34 & 5.33 & 153.36 & 2.94 & 196.84\\
                           & SEG    & 4.65 & 97.67 & 5.18 & 152.35 & 2.94 & 196.41\\
 \hline
  \multirow{4}{*}{StackGAN++} & TreeGAN& \textbf{3.90} & \textbf{101.59} & \textbf{4.33} & \textbf{145.44} & \textbf{3.10} &\textbf{186.45}\\
                              & No-PC  & 3.80 & 108.55 & 4.21 & 149.41 & 2.42 & 196.73\\
                              & No-SSR & 3.79 & 126.01 & 4.04 & 160.72 & 2.39 & 196.74\\
                              & PBG    & 3.71 & 115.26 & 3.98 & 155.37 & 2.41 & 204.01\\
                              & GloVe  & 3.82 & 124.95 & 4.14 & 162.81 & 2.40 & 194.85\\
                              & SEG    & 3.76 & 105.68 & 4.14 & 149.41 & 2.47 & 200.85\\
 \hline
   \multirow{4}{*}{GAN-INT-CLS} & TreeGAN& \textbf{3.87} & \textbf{85.82} & 4.03 & \textbf{112.76} & \textbf{2.23} & \textbf{85.86}\\
                                & No-PC  & 3.67 & 96.97 & 3.91 & 126.80 & 2.19 & 89.74 \\
                                & No-SSR & 3.48 & 121.73& 3.64 & 133.66 & 2.10 & 101.26\\  
                                & PBG    & 3.73 & 86.29 & 3.73 & 124.79 & 2.16 & 90.33\\  
                                & GloVe  & 3.65 & 87.76 & \textbf{4.06} & 129.54 & 2.12 & 94.57\\
                                & SEG    & 3.55 & 108.01 & 3.85 & 142.42 & 2.17 & 99.95\\
 \hline
\end{tabular}
 \label{tb:ablation}
\end{table*}

\begin{figure*}[htb]
	\centering
	\includegraphics[width=0.9\textwidth]{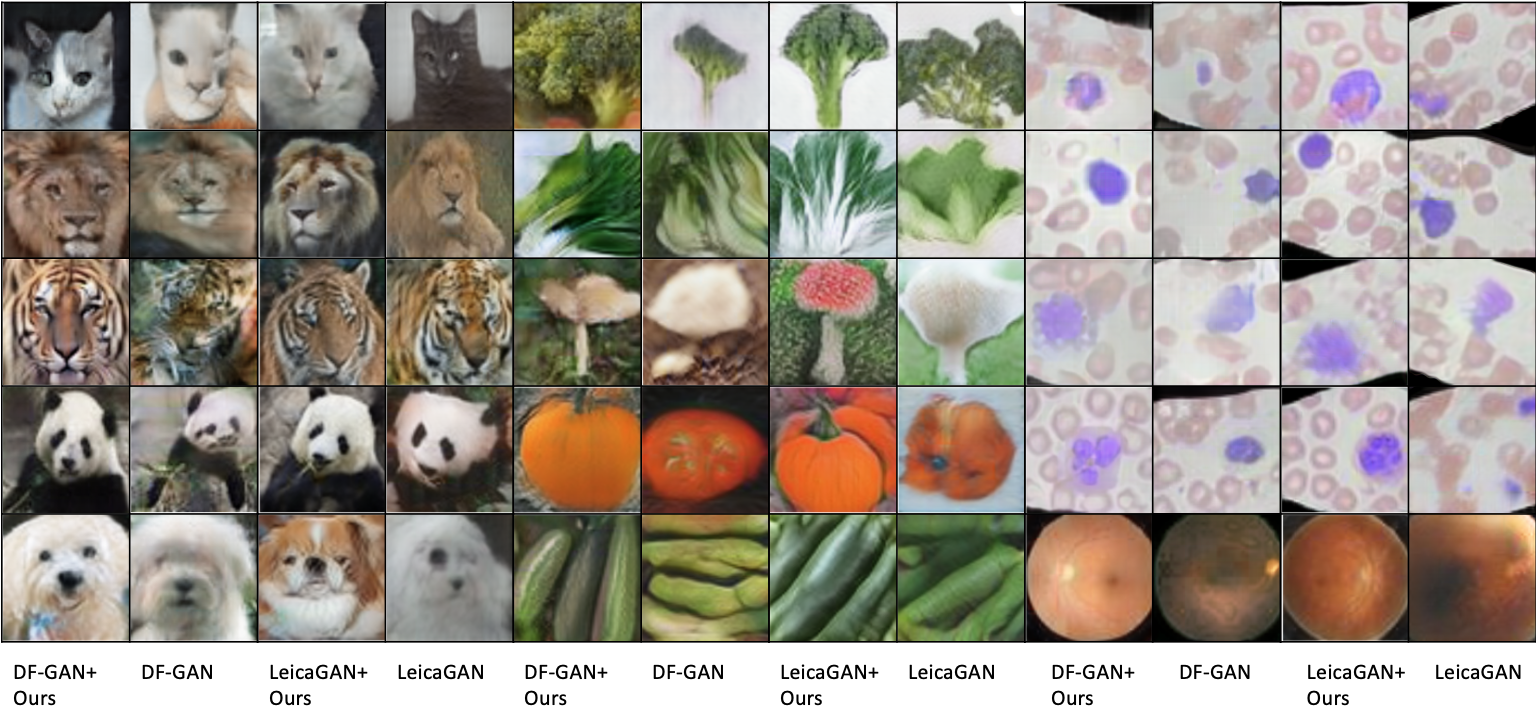}
	\caption{Images generated by our methods and baseline methods.}  
	\label{fig:imgs}
	\vspace{-0.3cm}
\end{figure*}

\subsection{Image Classification}
Table~\ref{tab:hp:hcc-acc2} shows the hierarchical classification accuracy on the training set, validation  set and test set of different datasets. The last column shows the difference between training accuracy and test accuracy. A smaller difference indicates less overfitting. As can be seen, in most cases, the train-test difference under self-supervised regularization (SSR) is smaller than that without using SSR. This indicates that SSR is an effective approach of alleviating overfitting.


\subsection{Class Hierarchies in the Three Datasets}
\begin{figure}[!t]
\centering
\subfloat[Animal]{\includegraphics[width=3.5in]{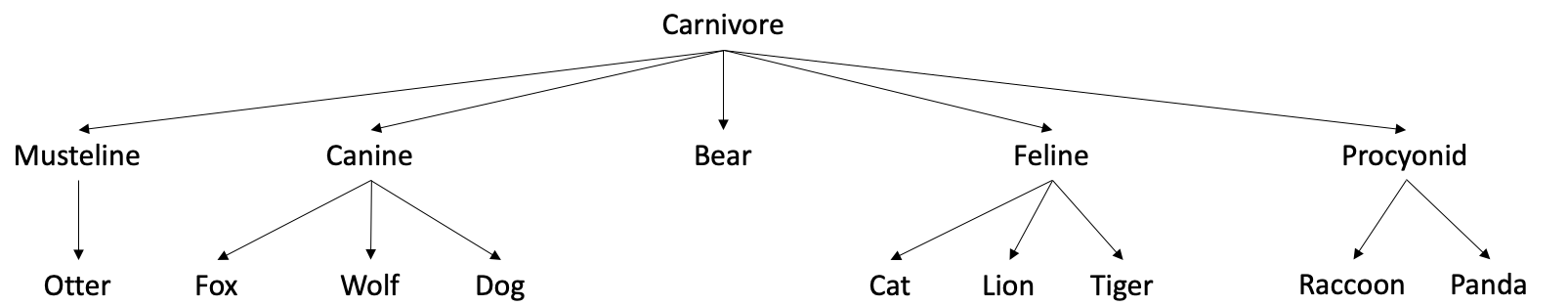}%
\label{fig_first_case}}
\hfil
\subfloat[Vegetable]{\includegraphics[width=3.5in]{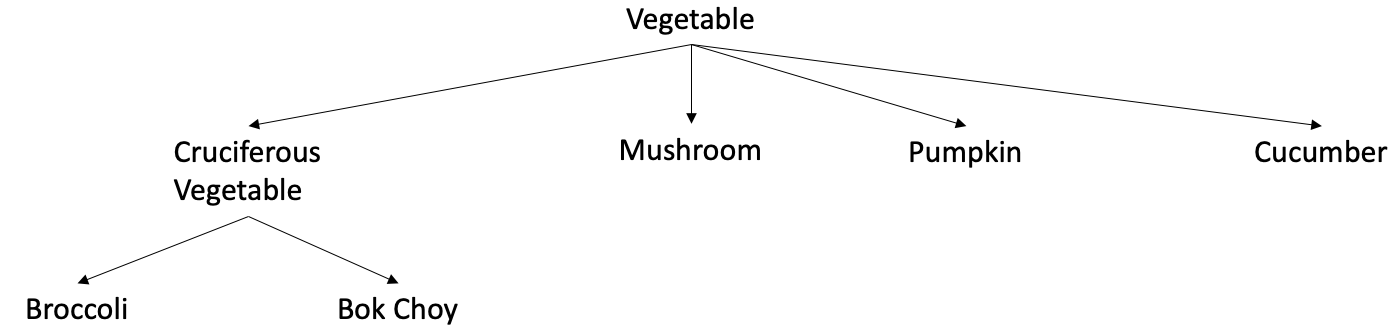}%
\label{fig_second_case}}
\hfil
\subfloat[Pathology]{\includegraphics[width=3.3in]{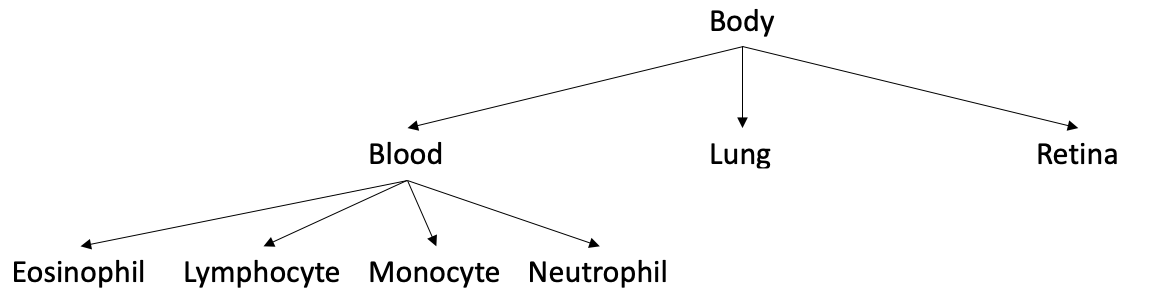}%
\label{fig_second_case}}
\caption{Image class hierarchy for animal, vegetable, and pathology classes}
\label{data}
\end{figure}

In this section, we show the class hierarchies of datasets used in the experiments. The animal class hierarchy, the vegetable class hierarchy and the pathology class hierarchy are shown in Figure~\ref{data}. From the figure, we can find that different classes share similar attributes in the class hierarchies, and one class can learn features from classes nearby. For example, dogs, wolves, and foxes are all under canine, and they share similar attributes: they all have shaggy hair and similar face attributes.


\subsection{Baselines}

We compare with the following baselines.

\begin{itemize}
 
\item \textbf{DF-GAN}~\cite{tao2020df}.  In this approach, the embeddings of class names are learned jointly with the embeddings of training images by minimizing a matching-aware gradient penalty. The class embeddings are fed into a conditional GAN to generate $256\times 256$ images. 


\item \textbf{LeicaGAN}~\cite{LeicaGAN} In this approach, the embeddings of class names are learned jointly with the embeddings of training images via a textual-visual co-embedding model. Then the class embeddings are fed into cascaded attentive generators to generate $256\times 256$ images.

\item \textbf{StackGAN++}~\cite{Han17stackgan2}.
In this approach, given a class name, it is encoded using a character-level CNN-RNN model, without considering the hierarchical relationship among classes.
Then the class encoding is fed into a stack of conditional GANs to generate  $256\times 256$ images. 

\item \textbf{GAN-INT-CLS}~\cite{reed2016generative}. In this approach, given a class name, it is encoded using a character-level CNN-RNN model, without considering the hierarchical relationship among classes. Then the class encoding is fed into a conditional GAN to generate $64\times 64$ images.  

\end{itemize}

\subsection{Results}

Table~\ref{tb:all} shows the IS and FID scores achieved by different methods on three datasets. In ``DF-GAN + Ours",  the class encoder in DF-GAN is replaced with our proposed class-hierarchy encoder; our proposed hierarchy-consistency checker is added to DF-GAN; the rest modules (especially the conditional image generator) are the same as DF-GAN. Similar setup is conducted in  ``LeicaGAN + Ours", ``StackGAN++ + Ours", and ``GAN-INT-CLS + Ours". From this table, we make the following observations. First,  ``DF-GAN + Ours" performs better than DF-GAN on all three datasets, with higher IS scores and lower FID scores. In DF-GAN, hierarchical relationships between classes are not incorporated. In contrast, our method leverages class hierarchy for image generation by learning hierarchy-aware class embeddings and performing posterior check to encourage generated images to be compatible with class hierarchy. This demonstrates that incorporating class hierarchy can indeed improve the fidelity of generated images and our proposed methods are effective in incorporating class hierarchy. Second, when our methods are applied to other conditional image generation (CIG) models including LeicaGAN, StackGAN++, and GAN-INT-CLS, the performance of these models are improved. This demonstrates that our methods are widely effective in improving various CIG models by incorporating the hierarchical relationship among classes into image generation.


\subsubsection{Ablation Studies}

We perform ablation studies to further verify the effectiveness  of individual modules in our model. We compare with the following ablation settings. Under each setting, one module is changed and the rest modules remain the same as those in the full TreeGAN model. 
\begin{itemize}
    \item \textbf{No post constraint (No-PC)}. Post constraint is not used. Generated images are not fed into the hierarchical classifier for consistency checking. The incorporation of class hierarchy is purely based on prior control.
    \item \textbf{No self-supervised regularization (No-SSR)}. In this setting, when training hierarchical classifiers in \ref{sec:hcc},  self-supervised regularization is not applied.
    \item \textbf{PBG}. In this setting, we replace our proposed log-determinant divergence (LDD) based embedding method with PBG~\cite{pbg}, an embedding method that takes hierarchical relationships between classes into account: the embeddings of a child and a parent are encouraged to be similar. 
    \item \textbf{GloVe}. In this setting, we replace our proposed log-determinant divergence (LDD) based embedding method with GloVe~\cite{glove}, an embedding method that does not take hierarchical relationships between classes into account. 
   \item  \textbf{Separate encoding and generation (SEG)}. In this setting, training of class-hierarchy encoder and training of conditional image generator are performed separately instead of jointly. We first learn  class embeddings offline, then use these fixed embeddings to train image generators. 
\end{itemize}

Table~\ref{tb:ablation} shows the results for different ablation studies. From this table, we make the following observations. \textbf{First}, TreeGAN  achieves higher (better) IS and lower (better) FID than No-PC. The only difference between TreeGAN and No-PC is that TreeGAN uses post constraint (PC) while No-PC does not. This demonstrates the effectiveness of post constraint in incorporating class hierarchy for generating higher-fidelity images. 
\textbf{Second}, TreeGAN  achieves higher (better) IS and lower (better) FID than No-SSR. The only difference between TreeGAN and No-SSR is that TreeGAN uses self-supervised regularization  during training hierarchical classifiers while No-SSR does not. This demonstrates the effectiveness of self-supervised regularization in alleviating overfitting of hierarchical classifiers, which enables image generators to produce more diverse images. Regarding alleviating overfitting, please see Table 7 in the supplements for more details. \textbf{Third}, TreeGAN performs better than PBG and GloVe, which demonstrates our proposed LDD-based embedding method is more effective than PBG and GloVe. GloVe does not consider class hierarchy, which hence is inferior. PBG only considers local and low-order hierarchical relationships between child-parent nodes while our method is able to incorporate long-range hierarchical relationships between any pair of nodes on the tree. Hence, our method yields better performance than PBG. \textbf{Fourth}, 
TreeGAN  achieves higher IS and lower FID than SEG. The only difference between TreeGAN and SEG is that TreeGAN learns  embeddings of class hierarchy and  image generator jointly while SEG performs that separately. This demonstrates the effectiveness of joint training, where the learning of a class embedding is guided by how well images in this class are generated.

\subsubsection{Qualitative Evaluation}

In addition to comparing different methods quantitatively, we also perform a qualitative evaluation by showing exemplar images generated by different methods. Figure~\ref{fig:imgs} shows some images generated by DF-GAN, ``DF-GAN + Ours", LeicaGAN, and ``LeicaGAN + Ours". 
From this figure, we can see that the images generated by ``DF-GAN + Ours" and ``LeicaGAN + Ours" which incorporate  hierarchical relationship among classes are more realistic, clear, and vivid than those generated by DF-GAN and LeicaGAN which do not consider class hierarchy. For example,  vegetable images generated by DF-GAN have a lot of blur or even are not recognizable. In contrast,  vegetable images generated by ``DF-GAN + Ours" are natural and realistic. This further demonstrates  leveraging class hierarchy can improve the fidelity of generated images.

\subsection{Visualization}

\begin{figure*}[h]
\centering
\includegraphics[width=0.85\textwidth]{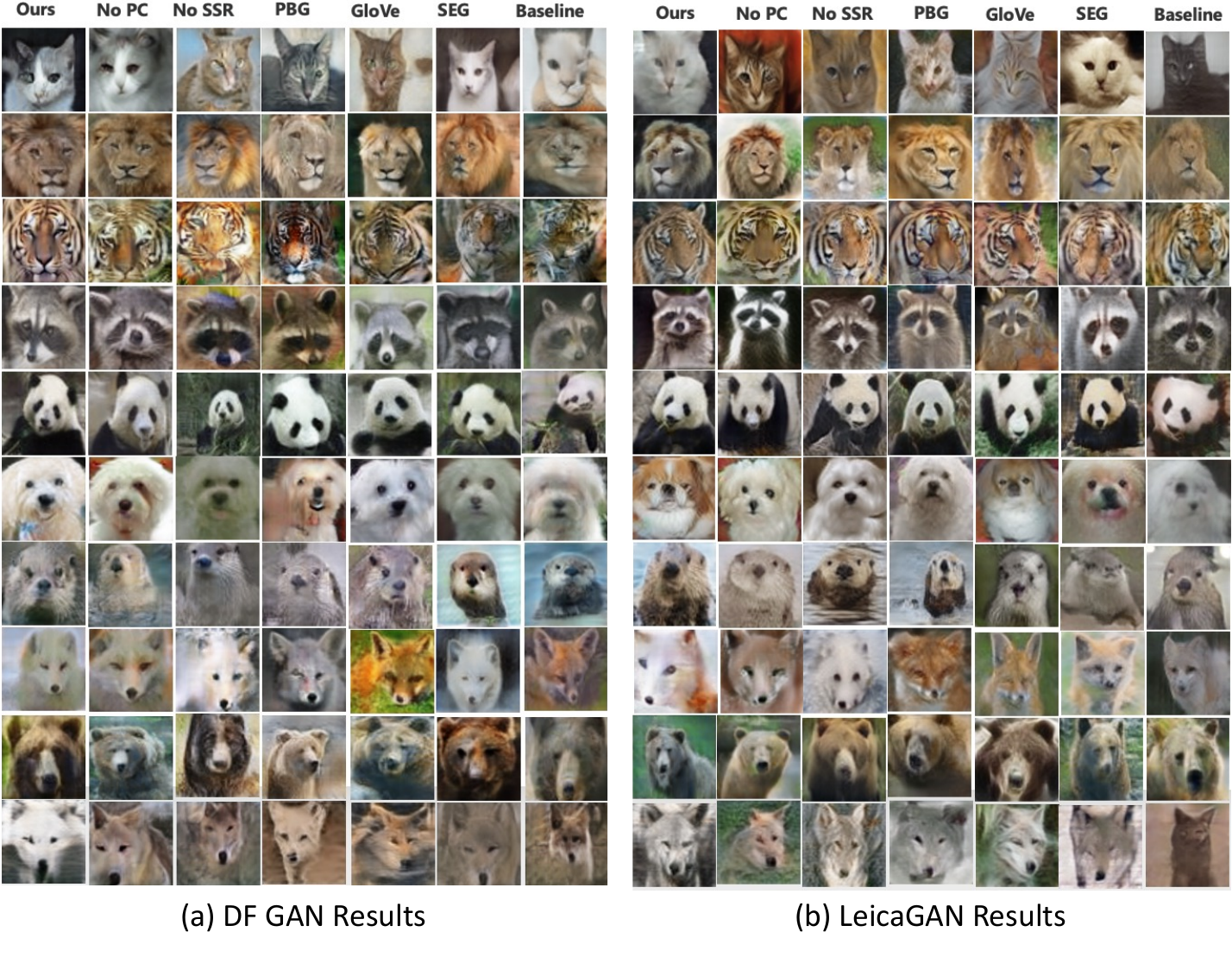}%

\caption{Generated animal images in the experiments}
\label{animal}
\end{figure*}

\begin{figure*}[h]
\centering
\includegraphics[width=0.9\textwidth]{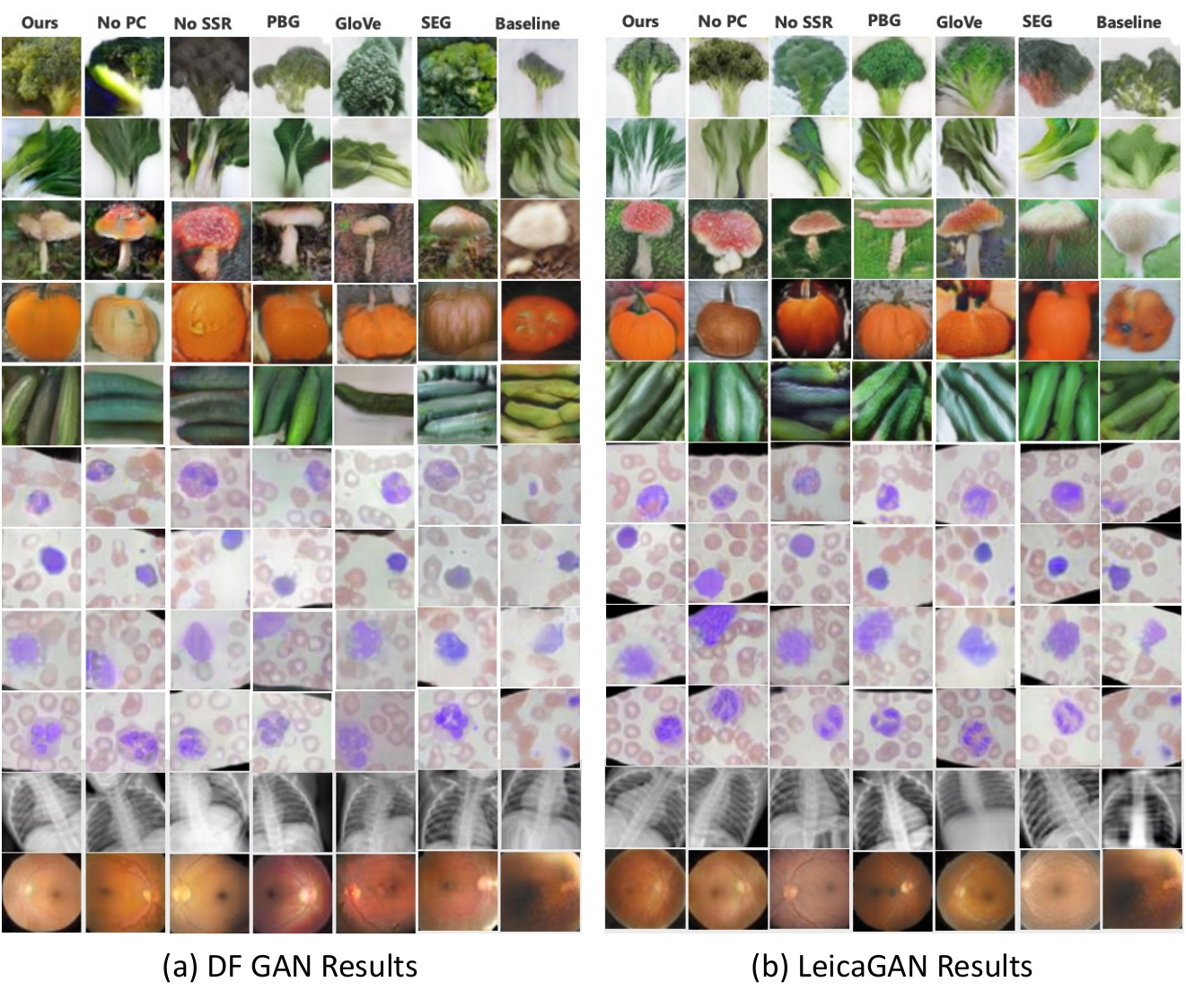}%

\caption{Generated vegetable and pathology images in the experiments}
\label{small}
\end{figure*}

The generated images by different methods, including TreeGAN, No-PC, No-SSR, PBG, GloVe, SEG, and baselines (DF-GAN, LeicaGAN, StackGAN++, and GAN-INT-CLS) are shown in Figure~\ref{animal} and Figure~\ref{small}. In Figure~\ref{animal}, we display the images with DF-GAN and LeicaGAN as the backbone. From left to right, we display the images generated using TreeGAN, No-PC, No-SSR, PBG, GloVe, SEG, and baselines in each subfigure. From up to bottom, we display the images belonging to cats, lions, tigers, raccoons, pandas, dogs, otters, foxes, bears, and wolves. In Figure~\ref{small}, we display the images with DF-GAN and LeicaGAN as the backbone. In each subfigure, from left to right, we display the images generated using TreeGAN, No-PC, No-SSR, PBG, GloVe, SEG, and baselines. From up to bottom, we first display the images that belong to the vegetable hierarchy, and the classes are broccoli, bok choy, mushroom, pumpkin, and cucumber. Then, we display the images that belong to the pathology hierarchy, and the classes are eosinophil, lymphocyte, monocyte, neutrophil, lung, and retina.

\section{Conclusions and Future Work}

In this paper, we propose to generate high-fidelity images by leveraging the hierarchical relationship among classes. To achieve this goal, we propose TreeGAN, which takes a hierarchy of classes as inputs and generates an image set for each class in the hierarchy. We propose two ways for incorporating a class hierarchy: prior control and post constraint. In prior control, the class hierarchy is encoded and fed into the generator to generate image sets. In post constraint, when the hierarchy of image sets are generated, we use an offline-trained classifier to check whether the generated images are consistent with the class hierarchy. The image generator is trained to maximize this consistency. We propose a novel hierarchy-aware embedding method and a novel regularization method based on self-supervised learning. 
Experiments on various datasets demonstrate the effectiveness of our proposed methods. For future works, we will leverage other types of class-relationship, such as graph relationship, for image generation. We will develop prior control and post constraint approaches that are tailored to graphs.

\bibliographystyle{IEEEtran}
\bibliography{reference}

\end{document}